\pdfoutput=1

\documentclass[11pt]{article}

\usepackage{EMNLP2023}

\usepackage{times}
\usepackage{latexsym}

\usepackage[T1]{fontenc}

\usepackage[utf8]{inputenc}

\usepackage{microtype}

\usepackage{inconsolata}
\usepackage{natbib}

\usepackage{multirow}
\usepackage{graphicx}
\usepackage{amssymb}
\usepackage{scalerel}
\usepackage[acronym]{glossaries}
\usepackage[abbreviations]{glossaries-extra}
\usepackage{xspace}
\usepackage{amsmath}
\usepackage{wasysym}
\usepackage{stackengine}
\usepackage{dsfont}
\usepackage{bbm}

\usepackage{subcaption}

\newcommand\quran{Qur’an\xspace}
\newcommand\KVHF{\stackon[-2pt]{\HF}{\HF}} 
\newcommand{\tablewidth}{\dimexpr(\textwidth-\columnsep)/2}

\newcommand\AarbA{\mbox{(ARB$^\circledcirc_\sim$)}\xspace}
\newcommand\AarbB{\mbox{(ARB$^\circledcirc_\approx$)}\xspace}

\newcommand\AelcA{\mbox{(ELC$^\otimes_\sim$)}\xspace}
\newcommand\AelcB{\mbox{(ELC$^\otimes_\approx$)}\xspace}

\newcommand\AcamB{\mbox{(CAM$^\otimes_\approx$)}\xspace}
\newcommand\AarbC{\mbox{(ARB$^\otimes_\sim$)}\xspace}
\newcommand\AarbD{\mbox{(ARB$^\otimes_\approx$)}\xspace}

\newcommand\elcQm{\mbox{(ELC$^\textit{M}_\sim$)}\xspace}

\newcommand\elcTf{\mbox{(ELC$^\textit{F}_\approx$)}\xspace}
\newcommand\elcTm{\mbox{(ELC$^\textit{M}_\approx$)}\xspace}

\newcommand\araTm{\mbox{(ARB$^\textit{M}_\approx$)}\xspace}

\newcommand\arabertbase{AraBERTv0.2-base\xspace}

\newcommand\camelCA{CAMeLBERT-CA\xspace}

\newcommand\elctra{AraELECTRA\xspace}

\usepackage{xcolor}
\usepackage{color}
\usepackage{todonotes}
\usepackage{comment}
\usepackage[normalem]{ulem}

\colorlet{kw}{blue}
\definecolor{com}{rgb}{0,0.6,0.3}

\newabbreviation{NLP}{NLP}{Natural language processing}
\newabbreviation{IR}{IR}{Information Retrieval}
\newabbreviation{QA}{QA}{Question Answering}
\newabbreviation{MRC}{MRC}{Machine Reading Comprehension}
\newabbreviation{MRS}{MRS}{Machine Reading at Scale}
\newabbreviation{CA}{CA}{Classical Arabic}
\newabbreviation{MSA}{MSA}{Modern standard Arabic}
\newabbreviation{DA}{DA}{Dialectal Arabic}
\newabbreviation{QRCDv1.2}{QRCDv1.2}{\quran reading comprehension dataset v1.2}
\newabbreviation{QRCDv1.1}{QRCDv1.1}{\quran reading comprehension dataset v1.1}
\newabbreviation{ARCD}{ARCD}{Arabic reading comprehension dataset}
\newabbreviation{AQQAC}{AQQAC}{Annotated Corpus of Arabic Al-\quran Question and Answer}
\newabbreviation{OSACT5}{OSACT5}{the $5^{th}$ Workshop on Open-Source Arabic Corpora and Processing Tools}
\newabbreviation{LM}{LM}{transformer-based pre-trained language models}
\newabbreviation{BERT}{BERT}{~\textbf{B}idirectional \textbf{E}ncoder \textbf{R}epresentations from \textbf{T}ransformers}
\newabbreviation{ELECTRA}{ELECTRA}{\textbf{E}fficiently \textbf{L}earning an \textbf{E}ncoder that \textbf{C}lassifies \textbf{T}oken \textbf{R}eplacements~\textbf{A}ccurately}
\newabbreviation{BIDAF}{BIDAF}{bidirectional attention flow for machine comprehension}
\newabbreviation{SOTA}{SOTA}{state-of-the-art}
\newabbreviation{MLM}{MLM}{masked language modeling}
\newabbreviation{T5}{T5}{Text-to-Text Transfer Transformer}
\newabbreviation{TANDA}{\textbf{T}AND\textbf{A}}{\textit{Transfer} and \textit{Adapt}}
\newabbreviation{EM}{EM}{Exact Match}
\newabbreviation{pRR}{pRR}{partial Reciprocal Rank}
\newabbreviation{MRR}{MRR}{mean Reciprocal Rank}
\newabbreviation{pAP}{pAP}{partial Average Precision }
\newabbreviation{MAP}{MAP}{mean Average Precision}

\newabbreviation{AP}{AP}{Average Precision}
\newabbreviation{RR}{RR}{Reciprocal Rank}
\newabbreviation{pP}{pP}{partial Precision}
\newabbreviation{HF}{HF}{Hugging Face}
\newabbreviation{MAL}{MAL}{Multi answer loss}
\newabbreviation{FAL}{FAL}{First answer loss}
\newabbreviation{TyDi}{TyDi QA}{Typologically Diverse Question Answering}
\newabbreviation{QPC}{QPC}{Thematic Qur'an Passage Collection}
\newabbreviation{STAR}{STAR}{\textbf{S}table \textbf{T}raining~\textbf{A}lgorithm for dense \textbf{R}etrieval}

\title{TCE at Qur'an QA 2023 Shared Task: Low Resource Enhanced Transformer-based Ensemble Approach for Qur'anic QA}

\author{Mohammed ElKomy, Amany Sarhan \\
Department of Computer Engineering, Faculty of Engineering  \\
Tanta University, Egypt \\
\texttt{\{mohammed.a.elkomy,amany\_sarhan\}@f-eng.tanta.edu.eg}
}

\begin{document}
    \maketitle
    \begin{abstract}
        In this paper, we present our approach to tackle \quran QA 2023 shared tasks~\textbf{A}~and~~\textbf{B}.
        To address the challenge of low-resourced training data, we rely on transfer learning together with
        a voting ensemble to improve prediction stability across multiple runs.
        Additionally, we employ different architectures and learning mechanisms for a range of Arabic pre-trained
        transformer-based models for both tasks.
        To identify unanswerable questions, we propose using a
        thresholding mechanism.
        Our top-performing systems greatly surpass the baseline performance on the
        hidden split, achieving a MAP score of 25.05\% for task
        ~\textbf{A} and a \gls*{pAP} of 57.11\% for task~\textbf{B}.


    \end{abstract}

    \section{Introduction}
    Ad hoc search is a fundamental task in \gls*{IR} and serves as the foundation for numerous \gls*{QA} systems and search engines.
    \gls*{MRC} is a long-standing endeavor in \gls*{NLP} and plays a significant role in the framework of text-based \gls*{QA} systems.
    The emergence of \gls*{BERT} and its family of \gls*{LM} have revolutionized the landscape of transfer learning systems
    for \gls*{NLP} and \gls*{IR} as a whole~\cite{ir_survey,quran_survey}.


    Arabic is widely spoken in the Middle East and North Africa, and among Muslims worldwide.
    Arabic is known for its extensive inflectional and derivational features.
    It has three main variants: \gls*{CA}, \gls*{MSA}, and \gls*{DA}.

    \quran QA 2023 shared task~\textbf{A} is a passage retrieval task organized to engage the community in conducting ad hoc search over the Holy \quran ~\cite{malhasphd,AyaTEC}.
    While \quran QA 2023 shared task~\textbf{B} is a ranking-based \gls*{MRC} over the Holy \quran, which is the second version of \quran QA 2022 shared task~\cite{overview2022,malhasphd}.

    This paper presents our approaches to solve the two tasks~\textbf{A} and~\textbf{B}.
    For task~\textbf{A}, 
    we explore both dual-encoders and cross-encoders for ad hoc search~\cite{ir_survey}.
    For task~\textbf{B}, we investigate \gls*{LM}s for extractive \gls*{QA} using two learning methods~\cite{bert}.
    For both tasks, we utilize various pre-trained Arabic \gls*{LM} variants.
    Moreover, we adopt external Arabic resources in our fine-tuning setups~\cite{malhasphd}.
    Finally, we employ an ensemble-based approach to account for inconsistencies among multiple runs.
    We contribute to the \gls*{NLP} community by releasing our experiment codes and trained \gls*{LM}s to GitHub~\footnote{\url{https://github.com/mohammed-elkomy/quran-qa}}.

    In this work, we address the following research questions \footnote{A superscript at the end of a RQ refers to one of the tasks. No superscript means the RQ applies for both tasks.}:\\
    \textbf{RQ1}: What is the impact of using external resources to perform pipelined fine-tuning?  \\
    \textbf{RQ2}: How does ensemble learning improve the performance obtained? \\
    \textbf{RQ3}: What is the effect of thresholding on zero-answer questions? \\
    \textbf{RQ4}\textsuperscript{A}: What is the impact of hard negatives on the dual-encoders approach?  \\
    \textbf{RQ5}\textsuperscript{B}: What is the impact of multi answer loss method on multi-answer cases?  \\
    \textbf{RQ6}\textsuperscript{B}: How is post-processing essential for ranking-based extractive question answering?

    The structure of our paper is as follows: Sections~\ref{sec:task-a} and
    \ref{sec:task-b} provide an overview of the datasets used in our study. In
    Section~\ref{sec:system}, we present the system design and implementation
    details for both tasks. The main results for both tasks are presented in
    Section~\ref{sec:results}. Section~\ref{sec:analysis} focuses on the analysis
    and discussion of our research questions \textbf{RQ}s. Finally, Section~\ref{sec:conc}
    concludes our work.

    \section{Task \textbf{A} Dataset Details}
    \label{sec:task-a}
    %
    %

    \begin{table}[]
        \centering
        \resizebox{0.9\columnwidth}{!}{%
            \begin{tabular}{cc|cc}
                \hline
                \multicolumn{2}{c|}{\textbf{Split}} & \textbf{Training} & \textbf{Development} \\ \hline
                \multicolumn{2}{c|}{\textbf{\begin{tabular}[c]{@{}c@{}}
                                                \# Question-passage\\ relevance pairs
                \end{tabular}}} & 972 & 160 \\ \hline
                \multicolumn{1}{c|}{\multirow{4}{*}{\textbf{\# Questions}}} & \textbf{Multi-answer} & 105 (60\%) & 15 (60\%) \\
                \multicolumn{1}{c|}{}                                       & \textbf{Single-answer} & \phantom{0}43 (25\%) & \phantom{0}6 (24\%) \\
                \multicolumn{1}{c|}{}                                       & \textbf{Zero-answer}   & \phantom{0}26 (15\%) & \phantom{0}4 (16\%) \\ \cline{2-4}
                \multicolumn{1}{c|}{}                                       & \textbf{Total}         & 174                  & 25                  \\ \hline
            \end{tabular}%
        }
        \caption{Task~\textbf{A} dataset relevance pairs distribution across training and development splits.
        We also include the distribution of answer types per split.}
        \label{tab:datastat-a}
    \end{table}

    \quran QA 2023 shared task~\textbf{A} serves as a test collection for the ad hoc
    retrieval task. The divine text is divided into segments known as the \gls*{QPC},
    where logical segments are formed based on common themes found among consecutive \quran{ic} verses~\cite{overview2023,Mushaf}.
    In this task, systems are required to provide responses to user
    questions in  \gls*{MSA} by retrieving relevant passages from
    the \gls*{QPC} when possible. This suggests there is a language gap between the questions and the passages, as the passages are in \gls*{CA}.
    Table ~\ref{tab:datastat-a} presents the distribution of the dataset across the training and development splits.
    The majority of questions in
    the dataset are multi-answer questions, meaning that systems can only receive
    full credit if they are able to identify all relevant passages for these
    queries.
    Additionally, Table ~\ref{tab:datastat-a} provides information on zero-answer questions,
    which are unanswerable questions from the entire \quran.
    (More information about the dataset distribution of topics in Appendix~\ref{appen:topics})

    Task~\textbf{A} is evaluated as a ranking task using the standard \gls*{MAP} metric.
    (Additional information about the evaluation process including zero-answers cases
    can be found in Appendix~\ref{appen:task-metric-a})

    \section{Task~\textbf{B} Dataset Details}
    \label{sec:task-b}

    \quran QA 2023 shared task~\textbf{B} is a ranking-based SQuADv2.0-like \gls*{MRC} over the Holy \quran, which extends to the \quran QA 2022~\cite{overview2022,squad}.
    The dataset is also referred to as \gls*{QRCDv1.2}.
    The same questions from task~\textbf{A} are organized as answer span extraction task from relevant passages~\cite{AyaTEC,overview2022}.
    (See the dataset distribution of topics in Appendix~\ref{appen:topics})

    \begin{table}[]
        \centering
        \resizebox{0.9\columnwidth}{!}{%
            \begin{tabular}{cc|cc}
                \hline
                \multicolumn{2}{c|}{\textbf{Split}} & \textbf{Training} & \textbf{Development} \\ \hline
                \multicolumn{2}{c|}{\textbf{\begin{tabular}[c]{@{}c@{}}
                                                \# Question-passage-answer \\ Triplets
                \end{tabular}}} & 1179 & 220 \\ \hline
                \multicolumn{1}{c|}{\multirow{4}{*}{\textbf{\begin{tabular}[c]{@{}c@{}}
                                                                \# Question-\\  passage\\ Pairs
                \end{tabular}}}} & \textbf{Multi-answer} & \phantom{9}134 (14\%)\phantom{0}                       & \phantom{11}29 (18\%)\phantom{0}                       \\
                \multicolumn{1}{c|}{} & \textbf{Single-answer}   & 806 (81\%)   & 124 (76\%)   \\
                \multicolumn{1}{c|}{} & \textbf{Zero-answer}         & \phantom{9}52 (5\%)\phantom{0}                              & \phantom{1}10 (6\%)\phantom{0}                              \\ \cline{2-4}
                \multicolumn{1}{c|}{} & \textbf{Total}         & 992                              & 163                              \\ \hline
            \end{tabular}%
        }
        \caption{Task~\textbf{B} dataset pairs and triplets distribution across training and development splits.
        For questions-passage pairs, we show the distribution of answer types.}
        \label{tab:datastat-b}
    \end{table}

    Table~\ref{tab:datastat-b} depicts the distribution of dataset pairs and triplets across the training and development splits.
    In addition, the table presents the distribution of answer types for the dataset pairs.
    \\
    Although zero-answer questions account for 15\% of the questions in task~\textbf{A} test collection,
    they only contribute to 5\% of the question-passage pairs in task~\textbf{B}.
    Furthermore, task~\textbf{B} has a limited number of unique questions in comparison to their corresponding question-passage pairs as seen from Tables~\ref{tab:datastat-a}~and~\ref{tab:datastat-b}, respectively.
    As a consequence, task~\textbf{B} can have repeated questions and passages among different samples
    and can be even \emph{leaked} among training and development splits~\cite{SMASH}.
    \citet{SMASH} analyzed this phenomenon and identified sources of \emph{leakage} in \gls*{QRCDv1.1}.
    In \gls*{QRCDv1.1}, leakage is defined as the presence of passages, questions,
    or answers that are shared among multiple samples~\cite{SMASH}. This can lead to \glspl*{LM}
    memorizing or overfitting leaked samples~\cite{SMASH}.
    \citet{SMASH} categorized
    \gls*{QRCDv1.1} into four distinct and mutually exclusive categories based on
    the type of leakage: pairs of passage-question, passage-answer, or just questions.
    (For more information about leakage in task~\textbf{B}, please refer to Appendix~\ref{appen:leakage})

    We extend the analysis made by~\citet{SMASH} for \gls*{QRCDv1.2}. Our main observation is that 90\% of the
    samples with no answer belong to the trivial leakage group called $D_{(1)}$.
    This group refers to samples with duplicate passage-answer or question-answer pairs.
    This indicates that zero-answer questions are not just less prevalent in task~\textbf{B} but also present a greater challenge in terms of generalization.
    Given the four groups defined by~\citet{SMASH}, they proposed a data re-splitting mechanism for \gls*{QRCDv1.1} called \emph{faithful} splits.
    In this work, we extend their resplitting approach and create faithful splits for \gls*{QRCDv1.2}.
    (Please refer to Appendix~\ref{appen:leakage} for more details about faithful splitting)

    Task~\textbf{B} is evaluated as a ranking task as well, using a recently proposed measure called
    \gls*{pAP}~\cite{AyaTEC,malhasphd}.
    (More details about this measure
    and zero-answer sample evaluation can be found in Appendix ~\ref{appen:task-metric-b})

    \section{System Design}
    \label{sec:system}

    \begin{table*}[]
        \centering
        \resizebox{1\textwidth}{!}{%
            \begin{tabular}{clcccccccccc}
                \hline
                \multicolumn{1}{c|}{\multirow{3}{*}{\textbf{\begin{tabular}[c]{@{}c@{}}
                                                                Short\\ Form
                \end{tabular}}}} & \multicolumn{1}{c|}{\multirow{3}{*}{\textbf{Systems}}} & \multicolumn{8}{c|}{\textbf{Single Model}} & \multicolumn{2}{c}{\textbf{Self Ensemble}} \\ \cline{3-12}
                \multicolumn{1}{c|}{} & \multicolumn{1}{c|}{} & \multirow{2}{*}{\textbf{MAP}} & \multirow{2}{*}{\textbf{MRR}}                      & \multirow{2}{*}{\textbf{R@10}}                      & \multirow{2}{*}{\textbf{R@100}}                      & \multicolumn{1}{c|}{\multirow{2}{*}{\textbf{\vspace{-3pt} MAP$^\bigstar_{\scaleto{\zeta}{10pt}}$}}}      & \multicolumn{3}{c|}{\textbf{MAP (Question Type)}}                                 & \multirow{2}{*}{\textbf{MAP}} & \multirow{2}{*}{\textbf{\vspace{-3pt} MAP$^\bigstar_{\scaleto{\zeta}{10pt}}$}} \\ \cline{8-10}
                \multicolumn{1}{c|}{}                                        & \multicolumn{1}{c|}{}     &                      &                      &                      &                      & \multicolumn{1}{c|}{}      & \textbf{Zero}        & \textbf{Single}      & \multicolumn{1}{c|}{\textbf{Multi}} &                               &                                                                                \\ \hline
                \multicolumn{1}{l}{Lexical Baseline}                         &                           & \multicolumn{1}{l}{} & \multicolumn{1}{l}{} & \multicolumn{1}{l}{}           & \multicolumn{1}{l}{}            & \multicolumn{1}{l}{}                                                                                & \multicolumn{1}{l}{} & \multicolumn{1}{l}{} & \multicolumn{1}{l}{}                &                               &                                                                                \\ \hline
                \multicolumn{1}{c|}{{\fontsize{10}{2}\selectfont BM$_\sim$}} & \multicolumn{1}{l|}{BM25}                           & 18.43 & 26.40 & 19.98                           & 19.98                            & \multicolumn{1}{c|}{26.40}                                                                           & 25.00                 & 16.67                 & \multicolumn{1}{c|}{17.39}           & N/A                          & N/A                                                                           \\ \hline
                \multicolumn{1}{l}{Dual-encoder}                             &                           & \multicolumn{1}{l}{} & \multicolumn{1}{l}{} & \multicolumn{1}{l}{} & \multicolumn{1}{l}{} & \multicolumn{1}{l}{} & \multicolumn{1}{l}{} & \multicolumn{1}{l}{} & \multicolumn{1}{l}{} & & \\ \hline
                \multicolumn{1}{c|}{{\fontsize{10}{2}\selectfont ARB$^\circledcirc_\sim$}} & \multicolumn{1}{l|}{
                    \begin{tabular}[c]{@{}l@{}}
                        \arabertbase\\ [-8.5pt] \phantom{xxx}$_{\textit{TASK A+ Random Neg}}$
                    \end{tabular}
                } & 20.02 & 42.87 & 29.72 & 48.23 & \multicolumn{1}{c|}{20.02} & 0.00 & 35.42 & \multicolumn{1}{c|}{19.20} & N/A & N/A \\
                \multicolumn{1}{c|}{{\fontsize{10}{2}\selectfont ARB$^\circledcirc_\approx$}} & \multicolumn{1}{l|}{
                    \begin{tabular}[c]{@{}l@{}}
                        \arabertbase\\ [-8.5pt] \phantom{xxx}$_{\textit{TASK A+ Hard Neg}}$
                    \end{tabular}
                } & \textbf{24.44} & 35.17 & 36.09 & 43.96 & \multicolumn{1}{c|}{24.44} & 0.00 & 45.00 & \multicolumn{1}{c|}{22.73}           & N/A                          & N/A                                                                           \\ \hline
                \multicolumn{1}{l}{Cross Encoder}                            &                           & \multicolumn{1}{l}{} & \multicolumn{1}{l}{} & \multicolumn{1}{l}{} & \multicolumn{1}{l}{} & \multicolumn{1}{l}{} & \multicolumn{1}{l}{} & \multicolumn{1}{l}{} & \multicolumn{1}{l}{} & & \\ \hline
                \multicolumn{1}{c|}{{\fontsize{10}{2}\selectfont ELC$^\otimes_\sim$}} & \multicolumn{1}{l|}{
                    \begin{tabular}[c]{@{}l@{}}
                        \elctra\\ [-8.5pt] \phantom{xxxxxxxxxx}$_{\textit{TASK A}}$
                    \end{tabular}
                } & 8.96 & 16.51 & 19.13 & 42.49 & \multicolumn{1}{c|}{16.48} & 3.00 & 10.32 & \multicolumn{1}{c|}{10.01} & 12.18 & 16.18 \\
                \multicolumn{1}{c|}{{\fontsize{10}{2}\selectfont ELC$^\otimes_\approx$}} & \multicolumn{1}{l|}{
                    \begin{tabular}[c]{@{}l@{}}
                        \elctra\\ [-8.5pt] \phantom{xx}$_{\textit{TyDi}\text{ }\textit{QA}_{\text{AR}}\rightarrow \textit{Tafseer} \rightarrow \textit{TASK A}}$
                    \end{tabular}
                } & 26.60 & 41.61 & 38.52 & 59.19 & \multicolumn{1}{c|}{31.91} & 19.00 & 38.31 & \multicolumn{1}{c|}{23.94} & 29.13 & 36.56 \\
                \multicolumn{1}{c|}{{\fontsize{10}{2}\selectfont CAM$^\otimes_\sim$}} & \multicolumn{1}{l|}{
                    \begin{tabular}[c]{@{}l@{}}
                        \camelCA\\ [-8.5pt] \phantom{xxxxxxxxxxx}$_{\textit{TASK A}}$
                    \end{tabular}
                } & 23.16 & 33.52 & 37.06 & 55.12 & \multicolumn{1}{c|}{27.45} & 13.00 & 36.92 & \multicolumn{1}{c|}{20.36} & 27.57 & 32.02 \\
                \multicolumn{1}{c|}{{\fontsize{10}{2}\selectfont CAM$^\otimes_\approx$}} & \multicolumn{1}{l|}{
                    \begin{tabular}[c]{@{}l@{}}
                        \camelCA\\ [-8.5pt] \phantom{xx}$_{\textit{TyDi}\text{ }\textit{QA}_{\text{AR}}\rightarrow \textit{Tafseer} \rightarrow \textit{TASK A}}$
                    \end{tabular}
                } & 29.34 & 42.17 & 39.93 & 57.23 & \multicolumn{1}{c|}{33.81} & 18.00 & \textbf{51.40} & \multicolumn{1}{c|}{23.54} & 32.77 & 36.77 \\
                \multicolumn{1}{c|}{{\fontsize{10}{2}\selectfont ARB$^\otimes_\sim$}} & \multicolumn{1}{l|}{
                    \begin{tabular}[c]{@{}l@{}}
                        \arabertbase\\ [-8.5pt] \phantom{xxxxxxxaxxxx}$_{\textit{TASK A}}$
                    \end{tabular}
                } & 31.76 & 41.93 & \textbf{46.55} & \textbf{62.71} & \multicolumn{1}{c|}{34.27} & \textbf{46.00} & 28.16 & \multicolumn{1}{c|}{\textbf{29.41}} & 36.09 & 36.87 \\
                \multicolumn{1}{c|}{{\fontsize{10}{2}\selectfont ARB$^\otimes_\approx$}} & \multicolumn{1}{l|}{
                    \begin{tabular}[c]{@{}l@{}}
                        \arabertbase\\ [-8.5pt] \phantom{xx}$_{\textit{TyDi}\text{ }\textit{QA}_{\text{AR}}\rightarrow \textit{Tafseer} \rightarrow \textit{TASK A}}$
                    \end{tabular}
                } & \textbf{34.83} & \textbf{47.09} & 39.99 & 60.82 & \multicolumn{1}{c|}{\textbf{37.55}} & 43.00 & 46.22 & \multicolumn{1}{c|}{28.10} & \textbf{36.70} & \textbf{40.70} \\ \hline
            \end{tabular}%
        }
        \caption{
            Dev split evaluation results for task~\textbf{A}. \textbf{MAP} means $\zeta$ is set to mark 15\% of questions as unanswerable. $\bigstar$~accompanied by $\zeta$ refers to applying the best $\zeta$ (see Appendix ~\ref{appen:optimal}).
            Average performance is reported for multiple runs of single models.
            Superscripts $\circledcirc$ and $\otimes$ in short form refer to dual-encoder and cross encoder, respectively.
            Subscripts $\sim$ and $\approx$ denote direct fine-tuning and pipelined fine-tuning, respectively.
        }
        \label{tab:dev-a}
    \end{table*}


    In this work, we fine-tune a variety of pre-trained Arabic \gls*{LM}s, namely
    \arabertbase~\cite{AraBERT}, \camelCA~\cite{arabic_models}, and \elctra~\cite{AraELECTRA}.
    We utilize transfer learning and ensemble learning for both tasks.
    To determine zero-answer cases, we apply a thresholding mechanism.
    (Additional information
    on transfer learning and ensemble learning can be found in Appendices~\ref{appen:transfer} and ~\ref{appen:ensemble}, respectively)

    \subsection{Task~\textbf{A} Architecture}

    We examine two distinct approaches for neural ranking in ad-hoc search:
    dual-encoders and cross-encoders approaches~\cite{ir_survey}.
%

    In dual-encoders, documents and queries are encoded separately into
    dense vectors, which are then compared using a metric learning function, such as
    cosine distance.
    We utilize \gls*{STAR} with a batch size of 16 queries to
    train our dense retrievers~\cite{star,ir_survey}.
    %

    In contrast cross-encoders involve encoding positive and negative
    pairs of documents and questions, assigning a relevance score.
    This method packs a document and a question into a single input for a sentence similarity \gls*{LM}~\cite{ir_survey}.
    Both methods require negative relevance signals during training.
    (Please refer to Figures~\ref{sfig:1}~and~\ref{sfig:2} in Appendix for both approaches.
    Additionally, see Appendix~\ref{appen:sys-a} for more details about negative selection criteria and zero-answer prediction)

    Although cross-encoders have a higher computational overhead compared to dual-encoders
    when used for ranking, the former has a quadratic complexity while the
    latter has a linear complexity. However, both methods are still feasible for
    low-resource datasets~\cite{ir_survey}.
    In both approaches, we utilize the cumulative predicted scores of the top K
    documents to calculate the likelihood of each question having an answer. We then
    apply a threshold $\zeta$ to identify zero-answer questions.

    \subsection{Task~\textbf{B} Architecture}
    We fine-tune pre-trained \gls*{LM}s for span prediction as in SQuADv2.0
    ~\cite{sqaud2,bert}.
    We use two different fine-tuning methods: \gls*{FAL} and \gls*{MAL}. The \gls*{FAL}
    method focuses on optimizing for the first answer in the ground truth answers,
    which is the default approach in standard span prediction implementations for SQuAD~\cite{bert,hugging}.
    In contrast, \gls*{MAL} optimizes for multiple answers simultaneously for the multi-answer samples in \gls*{QRCDv1.2}.
    This helps prevent the trained
    systems from being overly confident in a single span and distributes the
    predicted probability among different spans.
    (Refer to Appendix~\ref{appen:sys-b} for more information about these learning methods)

    It is worth noting that raw
    predictions from span prediction \gls*{LM}s are suboptimal for ranking \gls*{MRC},
    as many of them have overlapping content.
    To address this, we follow a post-processing mechanism proposed by~\citet{TCE}.
    (See Appendix~\ref{appen:imp-b} for implementation details)

    Similar to task~\textbf{A}, we perform thresholding by a hyperparameter $\zeta$ to determine
    zero-answer samples using \gls*{LM} null answer \textbf{[CLS]} token probability~\cite{sqaud2,bert}.
    (See Appendix~\ref{appen:threshold-b} for more details on zero-answer cases)

    \section{Results}
    \label{sec:results}

    The results tables for both tasks use the following notational format:
    We use short forms to refer to combinations of \glspl*{LM} and
    their fine-tuning approaches using superscripts and subscripts.

    The subscripts $\sim$ and $\approx$ denote direct fine-tuning and
    pipelined fine-tuning, respectively. Additionally, the arrows in model names
    subscripts indicate the stages of pipelined fine-tuning, with the learning resources names
    listed. Superscripts are used to denote the architectures employed for task~\textbf{A} and the learning methods for task~\textbf{B}.

    Tables~\ref{tab:dev-a} and~\ref{tab:dev-b} present our detailed results on
    the development split for both tasks for single and self-ensemble models.
    Table~\ref{tab:dev-a} shows the results for cross encoder and dual-encoders for task~\textbf{A}.
    Our best single model, \AarbD, achieved a \gls*{MAP} of 34.83\% and an MRR of 47.09\%.
    \AarbD self-ensemble achieved the best \gls*{MAP} of 36.70\%.
    Table~\ref{tab:dev-a} also presents the R@10 and R@100 metrics.
    This represents the upper bound on the reranking stage performance that we can obtain~\cite{ir_survey}.

    Table~\ref{tab:dev-b} summarizes the results for task~\textbf{B}.
    Our best performing model over the standard split,
    \elcTm, attained a \gls*{pAP} of 53.36\% and 55.21\% for single model and self-ensemble models, respectively.
    Table~\ref{tab:dev-b} also presents results for the faithful validation split we defined previously.
    \araTm is our best performing single model for the faithful split, achieving a \gls*{pAP} score of 54.19\%.


    \begin{table*}[]
        \centering
        \resizebox{1\textwidth}{!}{%
            \begin{tabular}{c|cc|ccccccccccc}
                \hline
                \multirow{4}{*}{\textbf{\begin{tabular}[c]{@{}c@{}}
                                            Short\\ Form
                \end{tabular}}} & \multicolumn{2}{c|}{\textbf{Systems}} & \multicolumn{8}{c|}{\textbf{Single Model}}                                                                                                                                                                                                                                                                                                                           & \multicolumn{3}{c}{\textbf{Self Ensemble Model}}                                                                                                                 \\ \cline{2-14}
                & \multicolumn{1}{c|}{\multirow{3}{*}{\textbf{Model}}} & \multirow{3}{*}{\textbf{Method}} & \multicolumn{2}{c|}{\textbf{Faithful}} & \multicolumn{9}{c}{\textbf{Standard Development Split}} \\ \cline{4-14}
                & \multicolumn{1}{c|}{} & & \multirow{2}{*}{\textbf{pAP}} & \multicolumn{1}{c|}{\multirow{2}{*}{\textbf{pAP$_{\textit{Post}}$}}}                           & \multirow{2}{*}{\textbf{pAP}}                            & \multirow{2}{*}{\textbf{pAP$_{\textit{Post}}$}}                            & \multicolumn{1}{c|}{\multirow{2}{*}{\textbf{\vspace{-3pt} pAP$^\bigstar_{\scaleto{\zeta}{10pt}}$}}}               & \multicolumn{3}{c|}{\textbf{pAP (Sample Type)}}                       & \multirow{2}{*}{\textbf{pAP}} & \multirow{2}{*}{\textbf{pAP$_{\textit{Post}}$}} & \multirow{2}{*}{\textbf{\vspace{-3pt} pAP$^\bigstar_{\scaleto{\zeta}{10pt}}$}} \\
                & \multicolumn{1}{c|}{} &     &                            & \multicolumn{1}{c|}{}                           &                            &                            & \multicolumn{1}{c|}{}               & \textbf{Zero} & \textbf{Single} & \multicolumn{1}{c|}{\textbf{Multi}} & & & \\ \hline
                {\fontsize{10}{2}\selectfont ELC$^\textit{F}_\sim$} & \multirow{2}{*}{
                    \begin{tabular}[c]{@{}l@{}}
                        \elctra\\ [-8.5pt] \phantom{xxxxxxxxxx}$_{\textit{TASK B}}$
                    \end{tabular}
                } & FAL & 34.97 & \multicolumn{1}{c|}{41.23} & 38.27 & 44.40 & \multicolumn{1}{c|}{39.26} & 18.67        & 41.51          & \multicolumn{1}{c|}{31.18}         & 41.16                        & 46.50                                          & 41.72                                                                         \\
                {\fontsize{10}{2}\selectfont ELC$^\textit{M}_\sim$}    &                       & MAL & \underline{37.44}          & \multicolumn{1}{c|}{\underline{42.63}} & \underline{40.55} & \underline{45.56} & \multicolumn{1}{c|}{41.48} & 14.67 & 43.69 & \multicolumn{1}{c|}{\underline{36.04}} & 42.01 & 47.21 & 43.90 \\ \hline
                {\fontsize{10}{2}\selectfont ELC$^\textit{F}_\approx$} & \multirow{2}{*}{
                    \begin{tabular}[c]{@{}l@{}}
                        \elctra\\ [-8.5pt] \phantom{xx}$_{\textit{TyDi}\text{ }\textit{QA}_{\text{AR}} \rightarrow \textit{TASK B}}$
                    \end{tabular}
                } & FAL & 52.76 & \multicolumn{1}{c|}{\underline{55.45}} & 49.76 & 53.70 & \multicolumn{1}{c|}{51.99}                                                                        & 10.33       & 54.36         & \multicolumn{1}{c|}{43.69}        & 50.66                       & 55.35                                         & 52.75                                                                        \\
                {\fontsize{10}{2}\selectfont ELC$^\textit{M}_\approx$} &                       & MAL & \underline{53.15}          & \multicolumn{1}{c|}{55.43} & \textbf{\underline{53.36}} & \textbf{\underline{56.42}} & \multicolumn{1}{c|}{\textbf{55.10}} & 18.33 & \textbf{56.61} & \multicolumn{1}{c|}{\textbf{\underline{51.55}}} & \textbf{55.21} & \textbf{58.38} & \textbf{57.05} \\ \hline
                {\fontsize{10}{2}\selectfont CAM$^\textit{F}_\sim$} & \multirow{2}{*}{
                    \begin{tabular}[c]{@{}l@{}}
                        \camelCA\\ [-8.5pt] \phantom{xxxxxxxxxxx}$_{\textit{TASK B}}$
                    \end{tabular}
                } & FAL & 41.45 & \multicolumn{1}{c|}{45.76} & 37.63 & 42.04 & \multicolumn{1}{c|}{38.36} & 11.00        & 40.83          & \multicolumn{1}{c|}{33.13}         & 42.51                        & 45.50                                          & 43.18                                                                         \\
                {\fontsize{10}{2}\selectfont CAM$^\textit{M}_\sim$}    &                       & MAL & \underline{43.54}          & \multicolumn{1}{c|}{\underline{47.36}} & \underline{38.57} & \underline{43.38} & \multicolumn{1}{c|}{39.38} & 12.67 & 40.52 & \multicolumn{1}{c|}{\underline{39.20}} & 41.66 & 45.39 & 43.80 \\ \hline
                {\fontsize{10}{2}\selectfont CAM$^\textit{F}_\approx$} & \multirow{2}{*}{
                    \begin{tabular}[c]{@{}l@{}}
                        \camelCA\\ [-8.5pt] \phantom{xxx}$_{\textit{TyDi}\text{ }\textit{QA}_{\text{AR}} \rightarrow \textit{TASK B}}$
                    \end{tabular}
                } & FAL & 50.64 & \multicolumn{1}{c|}{53.12} & \underline{41.59} & \underline{46.50} & \multicolumn{1}{c|}{42.39}                                                                        & 13.67       & 44.36         & \multicolumn{1}{c|}{39.39}        & 47.03                       & 49.37                                         & 47.12                                                                        \\
                {\fontsize{10}{2}\selectfont CAM$^\textit{M}_\approx$} &                       & MAL & \underline{52.14}          & \multicolumn{1}{c|}{\underline{54.01}} & 40.08 & 44.80 & \multicolumn{1}{c|}{41.30} & 15.00 & 41.61 & \multicolumn{1}{c|}{\underline{42.18}} & 42.75 & 46.87 & 44.23 \\ \hline
                {\fontsize{10}{2}\selectfont ARB$^\textit{F}_\sim$} & \multirow{2}{*}{
                    \begin{tabular}[c]{@{}l@{}}
                        \arabertbase\\ [-8.5pt] \phantom{xxxxxxxaxxxx}$_{\textit{TASK B}}$
                    \end{tabular}
                } & FAL & 44.81 & \multicolumn{1}{c|}{48.93} & 45.66 & \underline{49.34} & \multicolumn{1}{c|}{46.60}                                                                         & 23.67        & 49.29          & \multicolumn{1}{c|}{37.74}         & 49.38                        & 53.05                                          & 50.01                                                                         \\
                {\fontsize{10}{2}\selectfont ARB$^\textit{M}_\sim$}    &                       & MAL & \underline{47.41}          & \multicolumn{1}{c|}{\underline{50.62}} & \underline{45.71} & 47.69 & \multicolumn{1}{c|}{46.85} & 25.67 & 48.43 & \multicolumn{1}{c|}{\underline{41.03}} & 49.69 & 52.03 & 51.28 \\ \hline
                {\fontsize{10}{2}\selectfont ARB$^\textit{F}_\approx$} & \multirow{2}{*}{
                    \begin{tabular}[c]{@{}l@{}}
                        \arabertbase\\ [-8.5pt] \phantom{xxx}$_{\textit{TyDi}\text{ }\textit{QA}_{\text{AR}} \rightarrow \textit{TASK B}}$
                    \end{tabular}
                } & FAL & 52.97 & \multicolumn{1}{c|}{55.86} & \underline{50.62} & \underline{54.43} & \multicolumn{1}{c|}{51.28}                                                                        & \textbf{35.33}       & 53.78         & \multicolumn{1}{c|}{42.39}        & 52.20                       & 55.77                                         & 53.45                                                                        \\
                {\fontsize{10}{2}\selectfont ARB$^\textit{M}_\approx$} &                       & MAL & \textbf{\underline{54.19}} & \multicolumn{1}{c|}{\textbf{\underline{56.55}}} & 50.51                      & 53.32                      & \multicolumn{1}{c|}{51.35}          & 31.33         & 53.22           & \multicolumn{1}{c|}{\underline{45.54}}          & 52.13          & 54.94          & 52.94          \\ \hline
            \end{tabular}%
        }
        \caption{
            Dev split evaluation results for task \textbf{B}. \textbf{pAP} means fixing $\zeta$ to 0.8. \textbf{\textit{Post}} subscript identifies post-processing. $\bigstar$ accompanied by $\zeta$ refers to applying the best $\zeta$ (see Appendix ~\ref{appen:optimal}).
            Average performance is reported for multiple runs of single models.
            Superscripts F and M in short form indicate \gls*{FAL} and \gls*{MAL} methods, respectively.
            Subscripts $\sim$ and $\approx$ denote direct fine-tuning and pipelined fine-tuning, respectively.
            Underlined values refer to the higher performance when comparing the two learning methods.
        }
        \label{tab:dev-b}
    \end{table*}


    \begin{table}[]
        \centering
        \resizebox{0.842\columnwidth}{!}{%
            \begin{tabular}{cc|cc}
                \hline
                \multicolumn{1}{c|}{\textbf{\begin{tabular}[c]{@{}c@{}}
                                                Short\\ Form
                \end{tabular}}}                            & \textbf{Self Ensemble Model}                  & \textbf{MAP}   & \textbf{MRR} \\ \hline
                \multicolumn{2}{l|}{\textbf{TF-IDF Baseline}} & 9.03 & 22.60 \\ \hline
                \multicolumn{1}{c|}{CAM$^\otimes_\approx$} & \begin{tabular}[c]{@{}l@{}}
                                                                 \camelCA\\ [-8.5pt] \phantom{x}$_{\textit{TyDi}\text{ }\textit{QA}_{\text{AR}} \rightarrow  \textit{Tafseer}\rightarrow\textit{TASK A}}$     \\ [2pt]
                \end{tabular} & 23.02 & 47.06 \\
                \multicolumn{1}{c|}{ARB$^\otimes_\approx$} & \begin{tabular}[c]{@{}l@{}}
                                                                 \arabertbase\\ [-8.5pt] \phantom{x}$_{\textit{TyDi}\text{ }\textit{QA}_{\text{AR}} \rightarrow  \textit{Tafseer}\rightarrow\textit{TASK A}}$\\ [2pt]
                \end{tabular} & 24.64 & \textbf{49.39} \\
                \multicolumn{1}{c|}{MIX$^\otimes_\approx$} & CAM$^\otimes_\approx$ + ARB$^\otimes_\approx$ & \textbf{25.05} & 46.10        \\ \hline
            \end{tabular}%
        }
        \caption{Results on the hidden split for task~\textbf{A}.
            $\zeta$ is set to mark 15\% of questions as unanswerable. }
        \label{tab:test-results-a}
    \end{table}

    Both tables present comprehensive results for different question types, as well as
    the outcomes for a manually set threshold $\zeta$ and $\zeta^\bigstar$, i.e.,
    the threshold that yields the best performance. \\ \noindent(See Appendix~\ref{appen:optimal} for more details about $\zeta$ and optimal $\zeta$ selection)

    Considering the question types , experiments of \AarbC and \AarbD obtains the best \gls*{MAP} performance for zero-answer and multi-answer questions for task~\textbf{A}.

    With regard to the hidden split, Tables~\ref{tab:test-results-a} and~\ref{tab:test-results-b}
    provide a summary of our official submissions.

    In task~\textbf{A}, as
    shown in Table~\ref{tab:test-results-a}, we made 3 cross-encoder submissions:
    MIX$^\otimes_\approx$, which is an ensemble combining runs from
    CAM$^\otimes_\approx$ and ARB$^\otimes_\approx$ cross encoders. MIX$^\otimes_\approx$ achieved a \gls*{MAP} of
    25.05\%. In comparison, the TF-IDF baseline only achieved a \gls*{MAP} of 9.03\%.

    On the other hand, in task~\textbf{B}, we experimented with our two best performing models in Table~\ref{tab:dev-b}.
    As shown in Table~\ref{tab:test-results-b},
    \araTm outperformed \elcTm with a \gls*{pAP} of 57.11\%. This result
    is consistent with the findings from the faithful validation split~\cite{SMASH} in Table~\ref{tab:dev-b}
    for \araTm and \elcTm. Specifically, the
    \gls*{MAL} method outperformed \gls*{FAL} for all of our models in the faithful
    validation split (underlined in Table~\ref{tab:dev-b}).

    \section{Analysis and Discussion}
    \label{sec:analysis}

    Regarding \textbf{RQ1}, external resources always bring significant improvements to the same \gls*{LM} for
    both tasks.
    For task~\textbf{A}, we have three stages of fine-tuning as indicated by arrows in Table~\ref{tab:dev-a}.
    For example, when \AelcA is fine-tuned with external resources into \AelcB the \gls*{MAP} performance
    improves from 8.96\% to 26.60\% for single models as in Table~\ref{tab:dev-a}.
    In similar fashion for task~\textbf{B},
    \elcTm outperforms \elcQm by almost 13\% for the standard split in Table~\ref{tab:dev-b}.

    To answer our \textbf{RQ2}, ensemble learning consistently outperforms single models for both tasks.
    For instance, \AcamB ensemble surpasses its single model by 3.5\% for the \gls*{MAP} metric for task~\textbf{A}.
    Similarly, \elcTm ensemble outperforms its corresponding single model by almost a \gls*{pAP} of 2\% for task~\textbf{B}.

    With regard to \textbf{RQ3}, the hyperparameter $\zeta$ affects the zero answer type evaluation scores for both tasks.
    We make best use of the available data by employing a quantile method to determine the threshold $\zeta$ for both tasks.
    However, \AarbD model \gls*{MAP} performance improves by 3\% when the optimal $\zeta^\bigstar$ is employed for task~\textbf{A}.
    This suggests that there is a room for improvement for the $\zeta$ parameter.
    (Please refer to Appendix~\ref{appen:optimal} for more details about $\zeta$ selection and \textbf{RQ3}).

    In Table~\ref{tab:dev-a}, we experimented with dual-encoders using both random and hard negatives~\cite{star} to address \textbf{RQ4}.
    \AarbB outperforms \AarbA by almost 4.5\% when we perform hard negatives mining using a fine-tuned checkpoint \AarbA.

    In Table~\ref{tab:dev-b},
    \gls*{MAL} learning method consistently brings significant improvements to the final performance for all models over the faithful split.
    Moreover, it consistently outperforms \gls*{FAL} learning method for the multi-answer type of samples.
    For instance, \elcTm performs better than \elcTf, achieving a  \gls*{pAP} score of 51.55\%  compared to 43.69\% achieved by
    \elcTf for the subset of multi-answer samples.
    However, due to the fact that multi-answer samples make up only 18\%
    of the development samples in the standard split (Table~\ref{tab:datastat-b}),
    \gls*{MAL} does not always outperform \gls*{FAL} for the standard split overall performance.
    This finding addresses \textbf{RQ5}.

    With regard to \textbf{RQ6}, the post-processing approach proposed by~\citet{TCE} always surpasses the raw prediction score for both single and ensemble models.
    This is represented by \textbf{\textit{Post}} subscript in Table~\ref{tab:dev-b}.
    For example,  post-processing improves \araTm both single model and self-ensemble \gls*{pAP} performance by almost 3\%.
    \begin{table}[]
        \centering
        \resizebox{0.9\columnwidth}{!}{%
            \begin{tabular}{ccc|c}
                \hline
                \multicolumn{1}{c|}{\textbf{\begin{tabular}[c]{@{}c@{}}
                                                Short\\ Form
                \end{tabular}}}                                 & \multicolumn{1}{c|}{\textbf{Method}} & \textbf{Self Ensemble Model} & \textbf{pAP@10} \\ \hline
                \multicolumn{3}{l|}{\textbf{Full-passage Baseline}} & 32.68 \\ \hline
                \multicolumn{1}{c|}{ELC$^\textit{M}_\approx$} & \multicolumn{1}{c|}{\multirow{3}{*}{MAL}} & \begin{tabular}[c]{@{}l@{}}
                                                                                                                \elctra\\ [-8.5pt] \phantom{xx}$_{\textit{TyDi}\text{ }\textit{QA}_{\text{AR}} \rightarrow \textit{TASK B}}$
                                                                                                                \\ [2pt]
                \end{tabular} & 53.10 \\
                \multicolumn{1}{c|}{ARB$^\textit{M}_\approx$} & \multicolumn{1}{c|}{} & \begin{tabular}[c]{@{}l@{}}
                                                                                            \arabertbase\\ [-8.5pt] \phantom{xx}$_{\textit{TyDi}\text{ }\textit{QA}_{\text{AR}} \rightarrow \textit{TASK B}}$
                                                                                            \\ [2pt]
                \end{tabular} & \textbf{57.11}                \\
                \multicolumn{1}{c|}{MIX$^{\textit{M}}_\approx$} & \multicolumn{1}{c|}{}                & ELC$^\textit{M}_\approx$ + ARB$^\textit{M}_\approx$ & 56.43           \\ \hline
            \end{tabular}%
        }
        \caption{Results on the hidden split for task~\textbf{B}.
            $\zeta$ is set to mark 5\% of pairs as unanswerable. }
        \label{tab:test-results-b}
    \end{table}

    \section{Conclusion}
       \label{sec:conc}
    In this paper, we have presented our solution for both task~\textbf{A} and task~\textbf{B}  of  \quran  QA 2023 shared tasks.
    We explored various Arabic \gls*{LM}s
    using different training approaches and architectures.
    Our best performing
    systems are ensemble-based, enhanced with transfer learning using external learning
    resources.
    Lastly, we addressed a set of \textbf{RQ}s that highlight
    the main strengths of our work.

    \section*{Limitations}

    In this paper, we have adapted conventional learning-based architectures
    for Arabic \gls*{QA} tasks, specifically for \gls*{MRC} and ad hoc search.
    However, we faced several challenges
    throughout our study. One significant challenge was the scarcity of training
    resources, along with the imbalanced distribution of topics and question types. This
    was particularly evident in the zero-answer cases.
    As a consequence, our zero-answer thresholding mechanism demonstrated high sensitivity to each
    individual model.

    Additionally, we noticed significant performance variations
    due to the small size of the datasets.
    In order to tackle the problem of variations and noisy predictions, we
    investigated an ensemble approach. However, we still suggest that
    the results we obtained during the development phase may not accurately reflect
    the actual performance of learning systems.
    Despite the effectiveness of faithful splits for task~\textbf{B}, we still suggest exploring n-fold cross-validation for both tasks.
    However, our computation resources were significantly limited during the competition phase.


    For task~\textbf{B}, our models
    trained for \gls*{MRC} were found to be suboptimal for
    ranking tasks. Although our post-processing technique improved the raw
    predictions, this indicates the necessity for other ranking-based \gls*{MRC} approaches.
    Furthermore, we would like to explore the performance of large \gls*{LM}s on this particular task.

    \section*{Ethics Statement}
    The paper contains facts and beliefs that do not necessarily reflect
    the views or opinions of the authors. The information presented is based on
    objective analysis and does not aim to promote or endorse any particular
    religious interpretation.

    \section*{Acknowledgements}
    We would like to extend our heartfelt appreciation to Dr. Moustafa El Zantout
    for his invaluable support and insights during the course of this work.
    We would also like to express our deep gratitude to the organizers of
    \quran QA for their efforts in promoting research
    in Arabic in general, and the most significant Arabic text, the Holy \quran.

    \bibliography{anthology,custom}
    \bibliographystyle{acl_natbib}
    \section*{Appendix}
    \appendix

    \section{Dataset Additional Details}
\begin{figure*}[]%
        \center%
        \includegraphics[page=1,width=.88\textwidth]{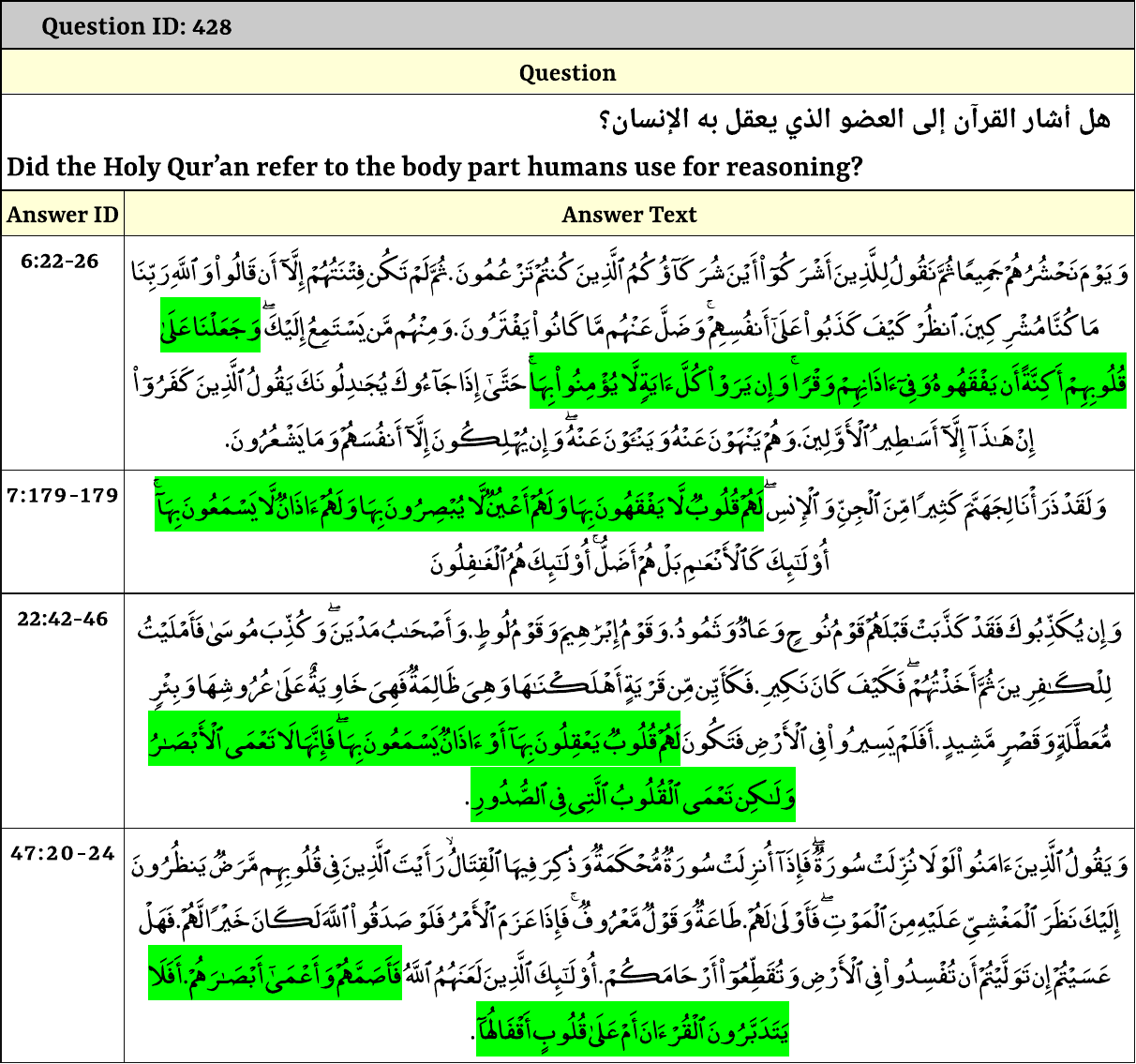}%
        \caption{A sample from shared task~\textbf{A}. We highlight the most relevant part in each \quran{ic} segment.}
        \label{fig:sample_a}
    \end{figure*}

\begin{figure*}[]%
        \center%
        \includegraphics[page=1,width=.88\textwidth]{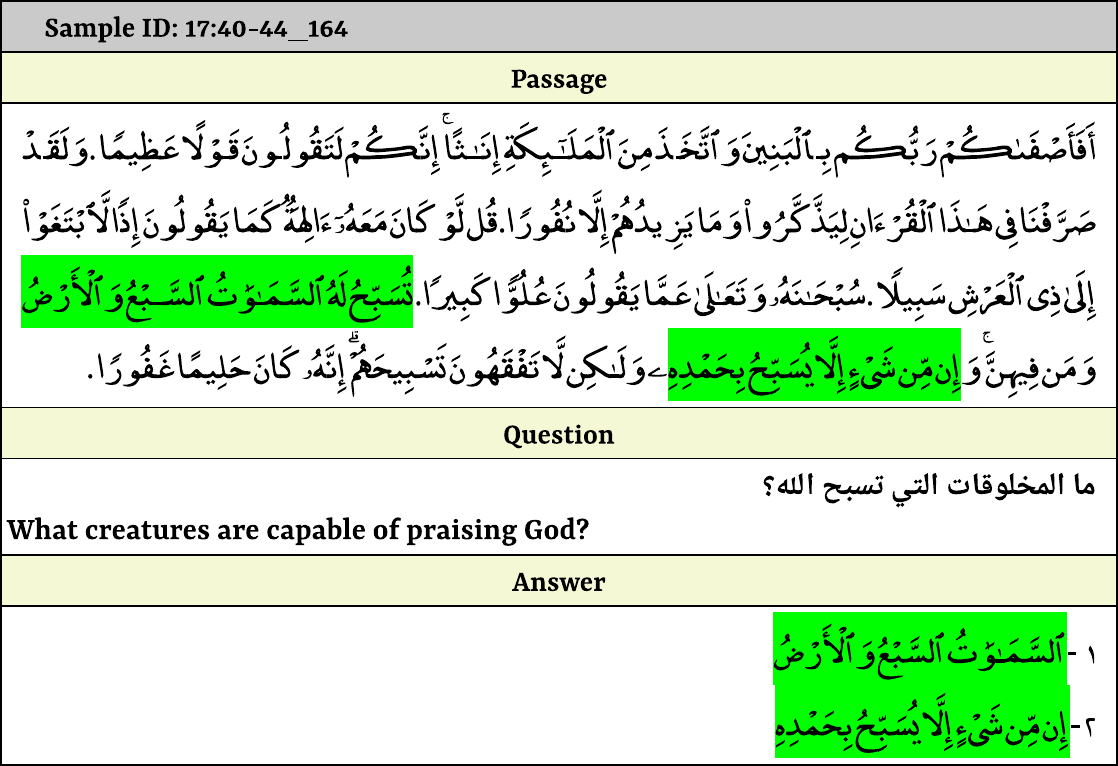}%
        \caption{A sample from shared task~\textbf{B}. We highlight the ground truth answers in the \quran{ic} passage.}
        \label{fig:sample_b}
    \end{figure*}

    AyaTEC is a dataset designed to evaluate the performance of retrieval-based
    Arabic QA systems over the Holy \quran.
    It contains 207 questions and 1,762 corresponding answers, which are categorized into 11 topics covering different aspects of the \quran.
    The dataset caters to the information needs of two types
    of users: skeptical and curious~\cite{AyaTEC}.
    The dataset includes single-answer and
    multi-answer questions, as well as questions that have no answer.
    Both \quran QA 2023 shared tasks are primarily based on an adapted version of AyaTEC~\cite{malhasphd,overview2022}.
    Figure~\ref{fig:sample_a} illustrates an example from task~\textbf{A}.
    The question asks whether there is a reference in the \quran to the body part used for reasoning.
    Four relevant \quran{ic} segments are annotated to have an answer for this question.
    Figure~\ref{fig:sample_b} depicts a question-passage-answer triplet from task~\textbf{B}.
    The question in this case is about creatures capable of praising God, within the context of the given passage.

    \subsection{Topic Distribution for tasks}
    \label{appen:topics}

\begin{figure*}[]%
        \center%
        \includegraphics[page=1,scale=.7]{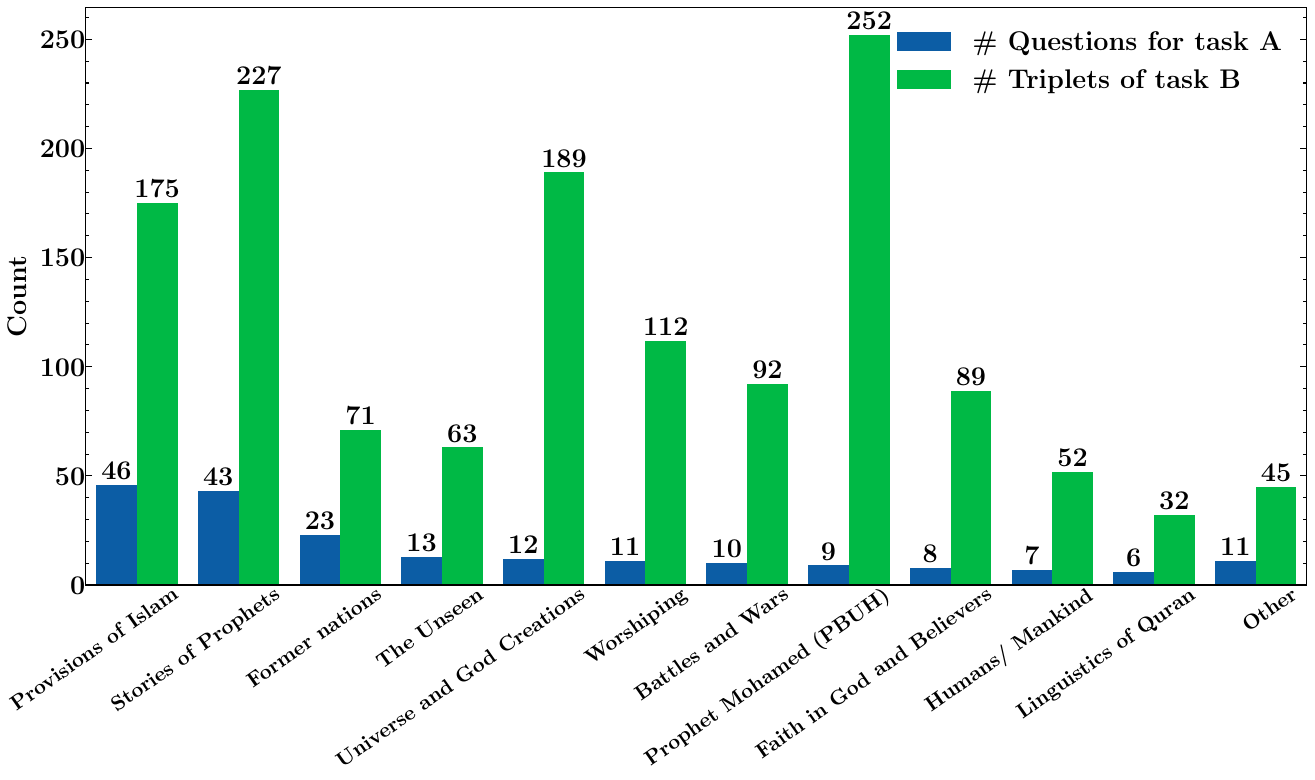}%
        \caption{Distribution of \gls*{QRCDv1.2} over the 11 topics for task~\textbf{A} questions and task~\textbf{B} triplets.}
        \label{fig:topic_dist}
    \end{figure*}

    AyaTEC covers 11 diverse topics referenced in the Holy \quran.
    Figure~\ref{fig:topic_dist} illustrates the imbalanced nature of those different topics.
    Furthermore, the representation of unique questions is significantly limited in comparison to question-passage-answer triplets.
    Additionally, it is evident that the ratio of triplets to unique questions varies for each respective topic.
    In task~\textbf{B}, these factors give rise to common questions across various passages.
    Consequently, they result in data leakage between the training and development splits~\cite{SMASH}.
    (Further information regarding this can be found in Appendix~\ref{appen:leakage})

    \subsection{Task~\textbf{A} Evaluation Measures}
    \label{appen:task-metric-a}
    For this ranking task, systems are expected to return up to 10 \quran{ic} passages for each question when possible.
    If the system determines that the question is unanswerable from the entire \quran, a null document is only returned, indicated by \hbox{-1}.
    The primary measure for the task is \gls*{MAP}, which gives full credit only if all relevant documents are retrieved at the top of the ranked answer list.
    For the zero-answer questions, full credit is given to successful systems
    only when they are unable to find any relevant \quran{ic} passage to answer the question, and return the null document.
    In addition to \gls*{MAP}, \gls*{MRR} is also reported, which gives credit just for the first relevant document from the ranked list~\cite{ir_survey}.

    In formal notation, we begin by defining the function  $\operatorname{\alpha}(q, p)$, which is a
    binary relevance function that indicates whether a passage $p$ is annotated as
    relevant to a question $q$ in the test collection. Equ.(\ref{mypsi}) represents the function that calculates
    the total number of relevant  \quran{ic} passages from the \gls*{QPC} to $q$.
    \begin{equation}
        \operatorname{\psi}(q) = {\sum_{p \in \mathcal{QPC}} \operatorname{\alpha}(q, p)}
        \label{mypsi}
    \end{equation}

    Zero-answer questions have a zero value for the function $\psi$, and their
    \gls*{MAP} score is calculated in a different way. Equ.(\ref{MAP}) shows the
    evaluation measure for \gls*{MAP} for answerable questions. For a
    ranked list $R$, we calculate the precision at each possible cutoff $@i$ at which a
    relevant document is present~\cite{ir_survey}.

    \begin{equation}
        \operatorname{MAP}(R, q)=\frac{\sum_{(i, p) \in R} \operatorname{Prec} @ i(R, q) \cdot \operatorname{\alpha}(q, p)}{ \operatorname{\psi}(q)},
        \label{MAP}
    \end{equation}

    Equ.(\ref{MAP_total}) illustrates the combined \gls*{MAP}
    evaluation measure for task~\textbf{A}. In this measure, zero-answer questions are given
    full credit only when $R$ is the null document,
    represented by $-1$ in the official evaluation script \footnote{ The symbol $\equiv$ signifies the equivalence operator between two lists.}~\cite{malhasphd}.

    \begin{equation}
        \operatorname{MAP_\text{A}}(R, q)=
        \begin{cases}
            \mathds{1}_{R \equiv [-1]} & \text{if } \psi(q) = 0 \\\\
            \operatorname{MAP}(R, q) &  \text{Otherwise}
        \end{cases}
        \label{MAP_total}
    \end{equation}
    $\mathds{1}_{C}$ is an \textit{indicator} function, which returns $1$ if the binary condition $C$ holds and $0$ otherwise.

    \subsection{Task~\textbf{B} Evaluation Measures}
    \label{appen:task-metric-b}
    Standard \gls*{MRC} tasks, like SQuADv2.0, are evaluated based only on the first prediction.
    In contrast, task~\textbf{B} is evaluated as a ranking task against a ranked list, rather than relying solely on the top prediction.
    As in task~\textbf{A}, systems are expected to return up to 10 answer spans from a given \quran{ic} passage to answer a question when possible.
    The primary evaluation metric for this task is \gls*{pAP}~\cite{AyaTEC,malhasphd}.
    This metric incorporates partial matching with the traditional rank-based Average Precision measure, i.e., \gls*{MAP}.
    In the case of unanswerable samples, the system receives a full score if it only returns and empty ranked list.

    Formally, partial matching is performed over token indexes of two substrings extracted from a given supporting passage.
    Based on~\citet{AyaTEC}, $F_{1}$ is used to calculate the similarity between the two substrings $R_{k}$ and $g$.
    $R_{k}$ represents the $k^{th}$ answer from a ranked list $R$, and $g$ refers to any ground truth answer from the set of ground truth answers $G$.
    \begin{equation}
        \mathcal{F_{\hspace{-2 pt}{\scaleto{1}{3pt}}}}^{R}_{k}  = \max_{g \in G}\left\{ F_{1}\left(R_{k},g\right)\right\}
        \label{eq:f1}
    \end{equation}
    In terms of Equ.(\ref{eq:f1}), we can define a partial matching version of precision at cutoff $K$, i.e., $\operatorname{pPrec}$~\cite{AyaTEC,malhasphd}.
    \begin{equation}
        \operatorname{pPrec}@K(R)=\frac{1}{K} \sum_{i=1}^K  \mathcal{F_{\hspace{-2 pt}{\scaleto{1}{3pt}}}}^{R}_{i}
        \label{eq:pPrec}
    \end{equation}
    In their study,~\citet{malhasphd} introduced a method for handling multi-answer
    samples. They proposed a string splitting mechanism that ensures only one
    correct answer is matched in each entry of $R$.
    Equ.(\ref{pAP}) presents the \gls*{pAP} evaluation metric for multi-answer ranking \gls*{MRC} in terms of $\operatorname{pPrec}$~\cite{cl_arabert},
    which stands as a token-level partial matching version of Equ(\ref{MAP}).
    \begin{equation}
        \operatorname{pAP}(R)=\frac{\sum_{i \in R} \operatorname{pPrec} @ i(R) \cdot \operatorname{\beta}(R,i)}{ |G|},
        \label{pAP}
    \end{equation}
    $\operatorname{\beta}(R,i)$ is a binary function that returns one if $R_i$ is a partially relevant answer.
    More specifically,
    \begin{equation}
        \operatorname{\beta}(R,k)=   \mathds{1}_{\mathcal{F_{\hspace{-2 pt}{\scaleto{1}{3pt}}}}^{R}_{k} > 0}
    \end{equation}

    In similar fashion, Equ.(\ref{pAP_total}) presents the complete \gls*{pAP}
    evaluation measure for task~\textbf{B}. In this measure, zero-answer samples are given
    full credit only when $R$ is an empty list~\cite{malhasphd}.

    \begin{equation}
        \operatorname{pAP_\text{B}}(R)=
        \begin{cases}
            \mathds{1}_{R \equiv [\text{ }]} & \text{if } |G| = 0 \\\\
            \operatorname{pAP}(R) &  \text{Otherwise}
        \end{cases}
        \label{pAP_total}
    \end{equation}

    \subsection{Leakage in \gls*{QRCDv1.2}}
    \label{appen:leakage}
    ~\citet{SMASH} analyzed \gls*{QRCDv1.1} and identified instances where passages and questions were repeated.
    They classified \gls*{QRCDv1.1} into four logical mutually-exclusive categories according to their complexity.
    Table~\ref{tab:smash} provides a summary of the criteria used and the expected behavior of
    trained \gls*{LM}s for each category.
    Additionally, symbols are employed to indicate the levels of complexity within each category, as determined by performance scores obtained by~\citet{SMASH}.
    Based on their analysis,~\citet{SMASH} solely utilized $D_{(3) \text { ood + hard }}$ for their final development split for \gls*{QRCDv1.1}.

    \begin{table*}[]
        \centering
        \resizebox{\textwidth}{!}{%
            \begin{tabular}{l|c|c}
                \hline
                \textbf{Category} & \textbf{Criteria} & \textbf{Expected \gls*{LM} behavior} \\ \hline
                $D_{(1) \text { in+leakage }} $ & \begin{tabular}[c]{@{}c@{}}
                                                      Samples with repeated passage-answer \\ or question-answer pairs
                \end{tabular} & \begin{tabular}[c]{@{}c@{}}
                                    Memorize answers and overfit to\\  training data $^{\AC}$
                \end{tabular} \\ \hline
                $D_{(2) \text { in}+\text {no leakage }}$ & \begin{tabular}[c]{@{}c@{}}
                                                                Samples with repeated passages but \\ having unique answers which are \\ different from  $D_{(1)}$ answers
                \end{tabular} & \begin{tabular}[c]{@{}c@{}}
                                    Reasoning is required to find the\\  right answer  \KVHF
                \end{tabular} \\ \hline
                $D_{(3) \text { ood + hard }}$ & \begin{tabular}[c]{@{}c@{}}
                                                     Samples with unique passages but \\ having rarely repeated questions \\ (appearing 3 times or less)
                \end{tabular} & \begin{tabular}[c]{@{}c@{}}
                                    Some reasoning is required to find \\ the right answer for rare questions $^{\VHF}$
                \end{tabular} \\ \hline
                $D_{(4) \text { ood + easy}}$ & \begin{tabular}[c]{@{}c@{}}
                                                    Samples with unique passages but \\ having commonly repeated questions \\ (more than 3 times)
                \end{tabular} & \begin{tabular}[c]{@{}c@{}}
                                    Lexical matching guides trained  \\ \gls*{LM}s to find similar answers  $^{\HF}$
                \end{tabular} \\ \hline
            \end{tabular}%
        }
        \caption{Description of the four categories introduced by~\citet{SMASH} over \gls*{QRCDv1.1} dataset.
        We show the criteria for identifying each category and the expected behavior for a fine-tuned \gls*{LM}.
        We denote the complexity of each category using symbols.
        For instance, \raisebox{-1pt}{\KVHF} represents the most challenging set for learning systems, while \AC \xspace refers the least challenging set.
        }
        \label{tab:smash}
    \end{table*}

    \begin{table*}[]

        \centering
        \resizebox{\textwidth}{!}{%
            \begin{tabular}{l|c|c}
                \hline
                \textbf{Category} & \textbf{Splitting Strategy by~\citet{SMASH}} & \textbf{Our Modified Splitting Strategy} \\ \hline
                $D_{(1) \text { in+leakage }}$ & \begin{tabular}[c]{@{}c@{}}
                                                     For duplicate question-answer or passage-answer pairs,\\ choose only one sample for training and leave the rest \\ for the development set.
                \end{tabular} & \begin{tabular}[c]{@{}c@{}}
                                    Use it entirely for training, this is due to the fact that\\  $D_{(1)}$ is trivial for development. \\ To balance the zero-answer questions ratio, we take entire\\  zero-answer leakage groups into the development set. \\ We employ disjoint-set algorithm for this purpose.
                \end{tabular} \\ \hline
                $D_{(2) \text { in}+\text {no leakage }}$ & \begin{tabular}[c]{@{}c@{}}
                                                                Split randomly with a splitting ratio of 86.7\% \\ for training and 13.3\% for development,\\  which corresponds to the original ratio of the data.
                \end{tabular} & \begin{tabular}[c]{@{}c@{}}
                                    Split them into two overlapping sets, as such, confusing examples\\  with the same passages are distributed among training and\\  development with different answers.
                \end{tabular} \\ \hline
                $D_{(3) \text { ood + hard }}$ & \begin{tabular}[c]{@{}c@{}}
                                                     Only use it for the development set \\ (removed from training).
                \end{tabular} & Same as~\citet{SMASH} \\ \hline
                $D_{(4) \text { ood + easy}}$ & \begin{tabular}[c]{@{}c@{}}
                                                    Split randomly with a splitting ratio of 86.7\% \\ for training and 13.3\% for development,\\  which corresponds to the original ratio of the data.
                \end{tabular} & \begin{tabular}[c]{@{}c@{}}
                                    Use it entirely for training, this is due to the fact that\\  $D_{(4)}$ is trivial for development.
                \end{tabular} \\ \hline
            \end{tabular}%
        }
        \caption{Description of our modified \textit{faithful} splitting for \gls*{QRCDv1.2} dataset over the four categories introduced by~\citet{SMASH}.
        We also show their proposed splitting approach~\cite{SMASH}.
        Check Table~\ref{tab:smash} for more details and reasons behind such splitting strategies.}
        \label{tab:smash_split}
    \end{table*}

    \begin{figure*}[]%
        \centering
        \begin{subfigure}{\columnwidth}
            \centering
            \includegraphics[height=135pt]{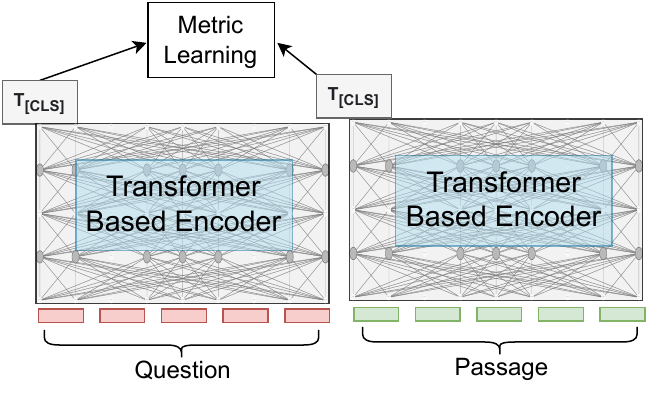}
            \caption{Dual-encoder generic architecture with metric learning for neural ranking.}
            \label{sfig:1}
        \end{subfigure} \hspace{10pt}
        \begin{subfigure}{\columnwidth}
            \centering
            \includegraphics[height=135pt]{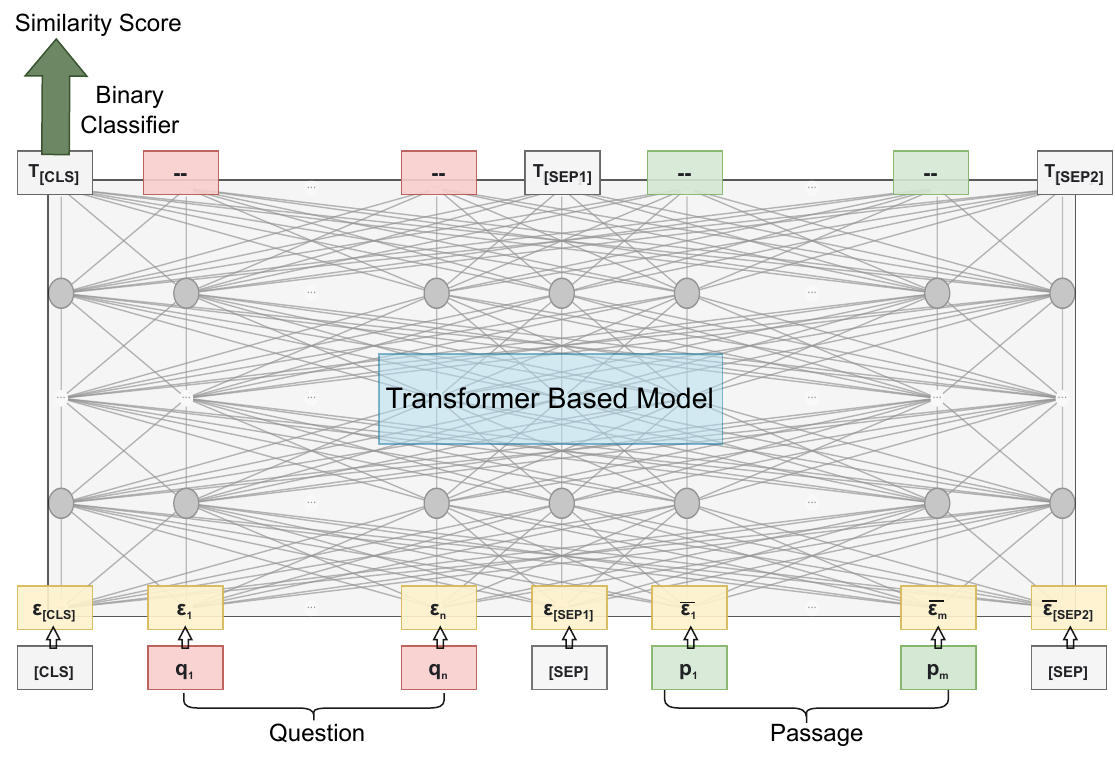}
            \caption{Cross-encoder generic architecture for an input pair of a question and a passage with a predicted similarity score.}
            \label{sfig:2}
        \end{subfigure}
        \caption{Diagrams for model architectures for task~\textbf{A}.}

    \end{figure*}

    In this work, we extend their approach for \gls*{QRCDv1.2}.
    We slightly modify this by considering both $D_{(2)}$ and $D_{(3)}$ for the development split.
    In addition, we employ disjoint set algorithm to find all leakage groups in $D_{(1)}$.
    We use those groups to balance the zero-answer questions ratio in the development split.
    This is because 90\% of zero-answer questions belong to the trivial leakage group $D_{(1)}$.

    In their work, \citet{SMASH} also proposed a resplitting approach for \gls*{QRCDv1.1}.
    They reorganized training and development splits using the four logical groups to create what they called \emph{faithful} splits for \gls*{QRCDv1.1}.
    Faithful splits aim to create more representative evaluations for \gls*{QRCDv1.1} dataset.
    Table~\ref{tab:smash_split} summarizes the modifications we made for performing evaluation using faithful splits.
    Table~\ref{tab:my_faith_dist} presents the distribution of
    our faithful split for \gls*{QRCDv1.2} based on our modified splitting strategy outlined in
    Table~\ref{tab:smash_split}. It also includes the distribution of zero-answer
    samples within each group.
    As in Table~\ref{tab:my_faith_dist}, we preserve the original ratio of training to development splits.
    Additionally, the percentage of zero-answer samples within each
    split is preserved compared to the original distribution in Table~\ref{tab:datastat-b}.

    \begin{table}[]
        \centering
        \resizebox{\tablewidth}{!}{%
            \begin{tabular}{l|cc|c}
                \hline
                \textbf{Category}                         & \textbf{Train}               & \textbf{Development}         & \textbf{Total}                \\ \hline
                $D_{(1) \text { in+leakage }}$            & 405 (49)                     & \phantom{00}7 (7)\phantom{1} & \phantom{1}412 (56)           \\
                $D_{(2) \text { in}+\text {no leakage }}$ & 290 (2)\phantom{1}           & \phantom{0}95 (1)\phantom{1} & \phantom{2}385 (3)\phantom{0} \\
                $D_{(3) \text { ood + hard }}$            & \phantom{00}0 (0)\phantom{5} & \phantom{0}62 (3)\phantom{1} & \phantom{12}62 (3)\phantom{0} \\
                $D_{(4) \text { ood + easy}}$             & 296 (0)\phantom{1}           & \phantom{00}0 (0)\phantom{1} & \phantom{1}296 (0)\phantom{0} \\ \hline
                Total                                     & 991 (51)                     & 164 (11)                     & 1155 (62)                     \\ \hline
                Zero-answer \%                            & 5.15 \%                      & 6.71 \%                      & 5.37 \%                       \\ \hline
            \end{tabular}%
        }
        \caption{\gls*{QRCDv1.2} dataset distribution of pairs for our \textit{faithful} splitting over the four categories introduced by~\cite{SMASH}.
        Parenthesized values refer to the number of zero-answer samples within each category for each split.}
        \label{tab:my_faith_dist}
    \end{table}

    \subsection{External Learning Resources}
    \label{appen:external}
    We leverage external resources to perform pipelined fine-tuning for both tasks
    ~\textbf{A} and~\textbf{B}.
    For task~\textbf{A}, we utilized interpretation resources (tafseer) from
    both Muyassar and Jalalayn, obtained from~\citet{Tanzil}.
    We created pairs of \gls*{QPC}
    \quran{ic} passages and their corresponding interpretations, resulting in
    approximately 2.5K relevant pairs.
    Additionally, we used the Arabic TyDI-QA Gold$\textbf{P}$ dataset~\cite{tydiqa} to generate pairs of relevant questions
    and their supporting evidence passages, resulting in 15K relevant pairs.
    For
    task~\textbf{B}, we solely relied on the Arabic subset of the TyDI-QA Gold$\textbf{P}$
    \gls*{MRC} dataset~\cite{tydiqa}. This dataset consists of
    approximately 15K question-passage-answer triplets.

    \section{Transfer Learning}
    \label{appen:transfer}
    In order to overcome the limited training resources for both tasks, we
    incorporate external \gls*{QA} and interpretation resources
    (tafseer)~\cite{Tanzil}.
    External resources enhance our learning systems in general by leveraging transfer learning across multiple fine-tuning stages~\cite{tanda,malhasphd}.
    We use arrows in subscripts in Tables~\ref{tab:dev-a},~\ref{tab:dev-b},~\ref{tab:test-results-a}, and~\ref{tab:test-results-b} to refer to stages of fine-tuning.
    (More details about external learning resources and
    their construction in Appendix~\ref{appen:external})

    \section{Ensemble Learning}
    \label{appen:ensemble}
    We utilize a voting self-ensemble technique for a group of fine-tuned models
    trained with different seeds~\cite{118}. We use the raw
    predictions without applying a zero-answer threshold.

    In task~\textbf{A}, for an ensemble $\mathcal{E}$ we aggregate the relevance
    scores for a \quran{ic} passage $p$ and a question $q$ assigned by a model $\varphi$.
    The ensemble relevance score $\mathcal{S}$ between $p$ and $q$ is as follows:
    \begin{equation}
        \operatorname{ \mathcal{S}}(q,p) = {\sum_{\varphi \in \mathcal{E}} \operatorname{\varphi}(q, p)}
        \label{ens-a}
    \end{equation}

    In similar fashion for task~\textbf{B}, we leverage a span voting ensemble~\cite{TCE}.
    For each sample, we aggregate span scores for each span $s$
    made by each predictor $\varphi$.
    \begin{equation}
        \operatorname{ \mathcal{S}}(s) = {\sum_{\varphi \in \mathcal{E}} \operatorname{\varphi}(s)}
        \label{ens-b}
    \end{equation}
    After that, we apply zero-answer thresholding to the aggregated result.

    \section{Additional System Details for task~\textbf{A}}
    \label{appen:sys-a}
    We summarize both architectures for task~\textbf{A} in Figures~\ref{sfig:1} and~\ref{sfig:2} for dual-encoders and cross-encoders, respectively.

    \subsection{Implementation Details}
    \label{appen:imp-a}
    In our  \gls*{STAR} training process, we incorporate both random in-batch negatives and
    hard negatives. Random negatives involve randomly selecting irrelevant documents
    for each query, providing positive and negative signals for learning systems~\cite{ir_survey}.
    On the other hand, hard negatives refer to the most offending
    irrelevant examples predicted by an encoder similarity score~\cite{star}. In a batch of size 16, we
    encode 16 different queries with their corresponding positive documents; in addition,
    in-batch negatives are used for all other queries. These negatives can be
    chosen randomly or through \gls*{STAR} hard negative mining.
    We use a learning rate of $5\times10^{-5}$ for all of our dual-encoder experiments.
    In the case of cross-encoders, we generate question-document pairs.
    These pairs have a ratio of one positive pair and three randomly selected negative pairs.
    For all of our cross-encoders, we use a learning rate of $1\times10^{-6}$ with a batch size of 16.

    \subsection{Zero-answer Prediction  }
    \label{appen:threshold-a}
    We assign a likelihood for each question $q$ to be answerable using
    the total relevance scores for the top returned passages $R$.
    $\varphi$ refers to a general relevance predictor between $q$ and a passage $p$.
    \begin{equation}
        \operatorname{\gamma}(q) = -{\sum_{p \in R} \operatorname{\varphi}(q, p)}
        \label{eq:threshold-a}
    \end{equation}
    The negative sign corresponds to the inverse proportional relationship between
    high relevance scores and the likelihood of unanswerability.
    We then normalize those scores for all questions into ${\operatorname{\bar{\gamma}}(q)}$ and apply
    a no answer threshold $\zeta$.
    We define a binary threshold function, $\operatorname{\sigma}$,
    which applies the threshold to identify unanswerable questions.
    \begin{equation}
        \operatorname{\sigma}(q)= \mathds{1}_{{\operatorname{\bar{\gamma}}(q)} > \zeta}
        \label{eq:threshold-a2}
    \end{equation}

    \section{Additional System Details for task~\textbf{B}}
        \label{appen:sys-b}
\begin{figure*}[]%
        \center%
        \includegraphics[page=1,scale=.7]{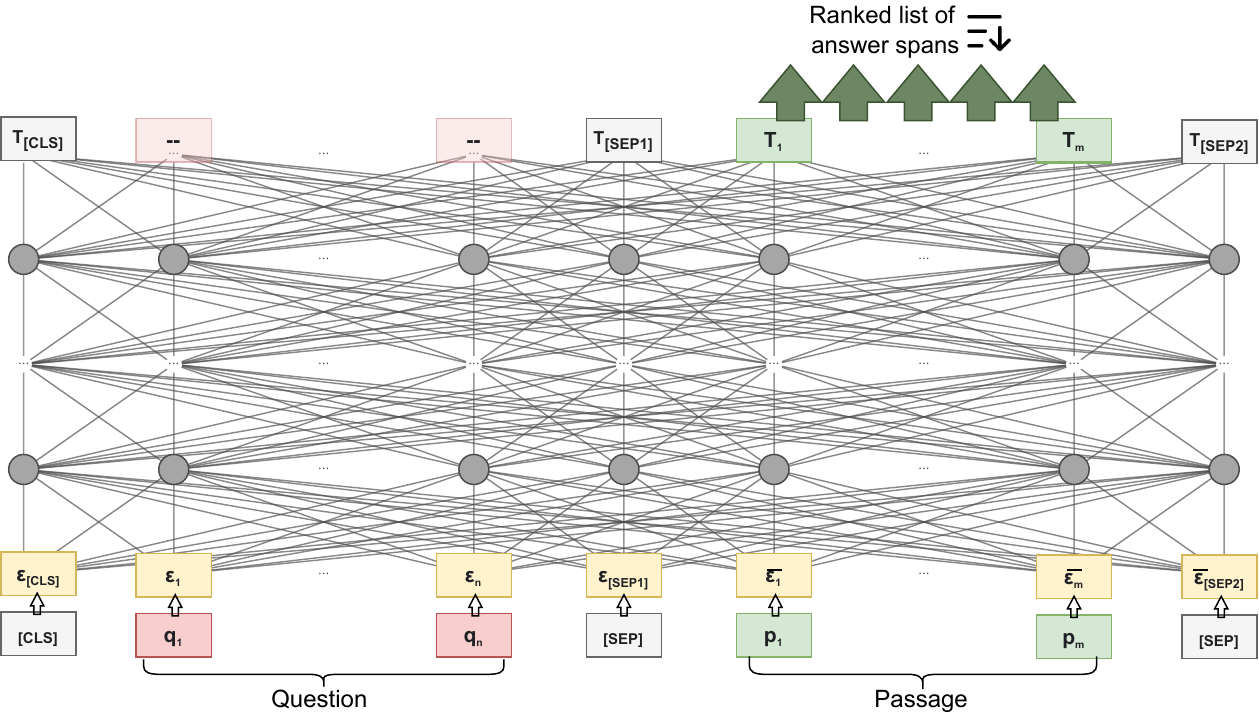}%
        \caption{Generic architecture illustration of a \gls*{LM} for ranking \gls*{MRC}.}
        \label{fig:bertQA}
    \end{figure*}

    In this work, we fine-tune \gls*{LM}s for extractive \gls*{MRC} as span predictors~\cite{bert}.
    The fine-tuning process involves packing each
    question-passage pair $x$ together and feeding it to a \gls*{LM} to predict the start and end token indices from the passage, as shown in Figure~\ref{fig:bertQA}.
    To achieve this, a trainable randomly
    initialized start vector $S$ and end vector $E$ are stacked on top of the  \gls*{LM},
    having the $i^{th}$ token hidden-representation $T_{i}$.
    The final model with
    the newly stacked layers has learnable parameters $\theta$.

    \begin{figure*}[]%
        \centering
        \begin{subfigure}{\columnwidth}
            \centering
            \includegraphics[width=0.8\columnwidth,height=5cm]{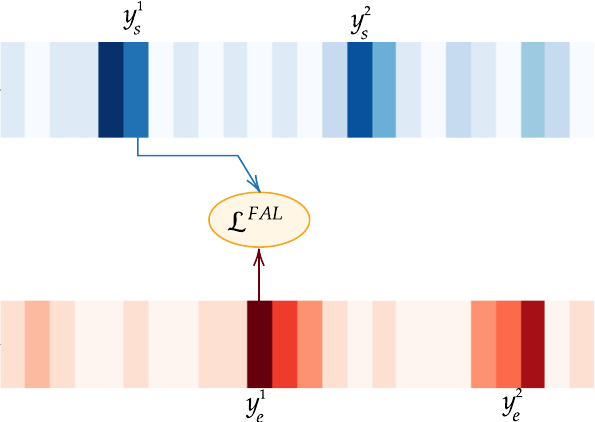}
            \caption{Standard (\gls*{FAL})}
            \label{sfig:standard}
        \end{subfigure} \hspace{5pt}
        \begin{subfigure}{\columnwidth}
            \centering
            \includegraphics[width=0.8\columnwidth,height=5cm]{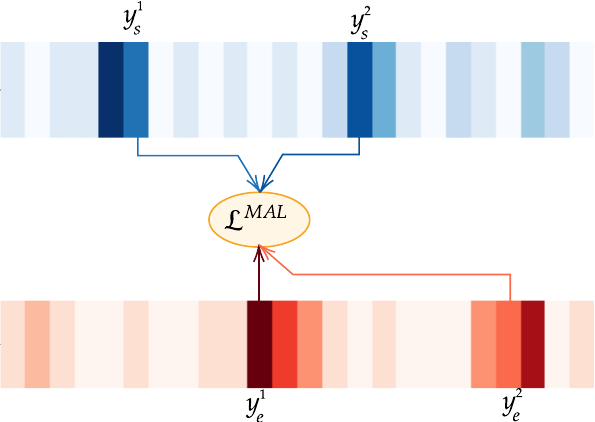}
            \caption{\gls*{MAL}}
            \label{sfig:MAL}
        \end{subfigure}
        \caption{Illustration of Learning Methods.}
        \label{fig:learning}
    \end{figure*}

    \begin{figure*}[]%
        \center%
        \includegraphics[page=1,width=.7\textwidth]{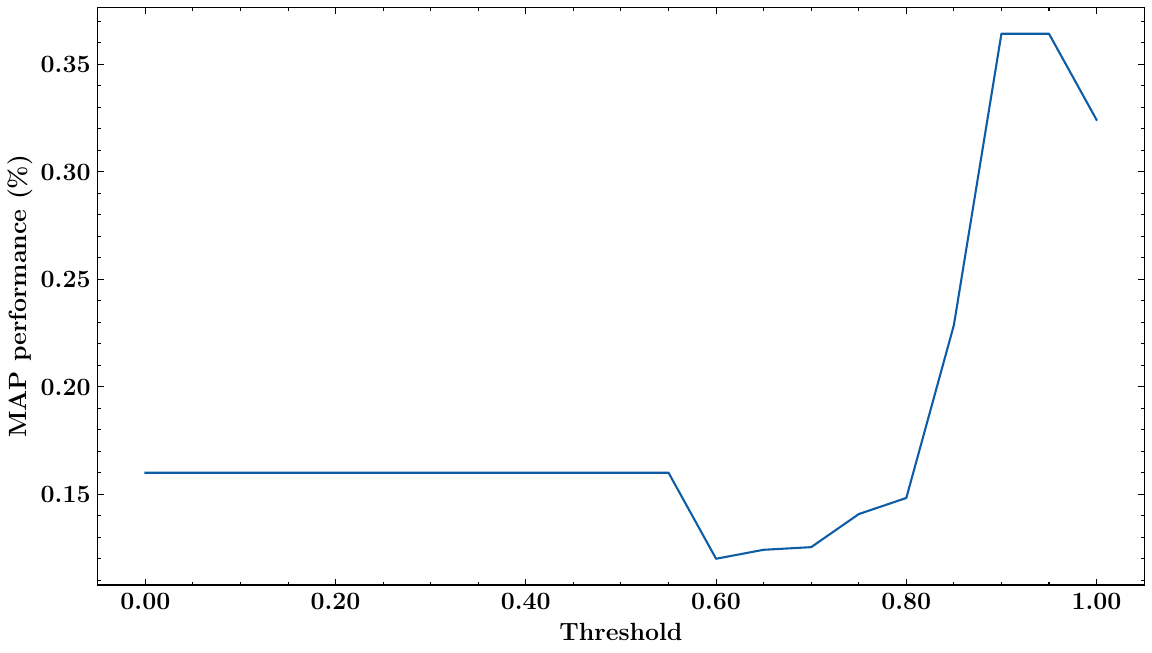}%
        \caption{Thresholding effect against \gls*{MAP} performance for one of our fine-tuned models.}
        \label{fig:threshold}
    \end{figure*}

    The dot product between $S$ and $T_{i}$ is chosen to determine the score that the $i^{th}$ token is the start of the answer span.
    These scores for all passage tokens are
    followed by a softmax layer that produces the probabilities for individual
    tokens being the start of the answer span~\cite{BIDAF,bert}.
    Equ.(\ref{eqn:start_softmax})
    depicts the probability that the $i^{th}$ token is the start of the answer span.

    \begin{equation}
        \mathbb{P}\left(i \mid x ; \theta\right)=\frac{e^{S \cdot T_{i}}}{\sum_{j}^{|T|} e^{S \cdot T_{j}}}
        \label{eqn:start_softmax}
    \end{equation}

    Under full-supervision, the training objective is to optimize the log-likelihoods for both the ground truth start and end positions.
    For a model with learnable $\theta$, an input $x$, and a single ground truth answer span $y$, the log likelihood for the start token position is as follows:
    \begin{equation}
        \mathcal{L}_{\text {start}}\left(\theta \mid x, y\right)=-\log \mathbb{P} \left({y^{}_{_{\scaleto{s}{3pt}}}} {\mid x ; \theta}\right)
        \label{eqn:loss_single}
    \end{equation}
    where the subscript $s$ in ${y^{}_{_{\scaleto{s}{3pt}}}}$ refers to the start position of the answer span $y$.

    If there are multiple answers for a sample $x$, we rather have a set of plausible answer spans $\mathcal{Y}$.
    ~\citet{TCE,Stars,gof} in \quran QA 2022 tackled this by considering any answer span from $\mathcal{Y}$ by taking one at random or the first answer span, namely, $y^1$.
    We denote the $i^{th}$ answer from $\mathcal{Y}$ as $y^i$.
    We call this learning method First answer loss (FAL). This can be formulated in terms of $\mathcal{Y}$ as denoted below:
    \begin{equation}
        \mathcal{L}\strut^{\text {FAL}}_{\text {start}}\left(\theta \mid x, \mathcal{Y}\right)=-\log \mathbb{P}\left({y^1_{_{\scaleto{s}{3pt}}}} \mid x ; \theta\right)
        \label{eqn:loss_fal}
    \end{equation}
    Figure~\ref{sfig:standard} illustrates this learning method.
    However, \gls*{QRCDv1.2} task \textbf{B} considers a multi-answer \gls*{MRC} scenario, this leads to discrepancy between
    training and testing when \gls*{FAL} learning method is employed for fine-tuning.
    Towards this end, we define \gls*{MAL} learning method. This learning method takes the multi-answer cases in consideration by
    optimizing for all answers altogether.
    Mathematically, this generalizes to any $y^i$ from the set $\mathcal{Y}$ and takes the sum of the log likelihood losses for multiple answers
    as shown in Equ.(\ref{eqn:loss_mal}):
    \begin{equation}
        \mathcal{L}\strut^{\text {MAL}}_{\text {start}}\left(\theta \mid x, \mathcal{Y}\right)=-\sum_{y^i \in \mathcal{Y}}{\log \mathbb{P}\left({y^i_{_{\scaleto{s}{3pt}}}} \mid x ; \theta\right)}
        \label{eqn:loss_mal}
    \end{equation}

    We show the \gls*{MAL} learning method in Figure~\ref{sfig:MAL}.

    \subsection{Implementation Details}
    \label{appen:imp-b}
    To enhance \glspl*{LM} predictions, we employ a post-processing approach.
    ~\citet{TCE} proposed an effective non-maximum suppression post-processing approach at \quran QA 2022~\cite{overview2022}.
    They also proposed some operations for rejecting uninformative short answers.
    For all of our models, we used a learning rate of $2\times10^{-5}$ and a batch size of 16.

    \subsection{Zero-answer Prediction  }
    \label{appen:threshold-b}
    \gls*{MRC} for SQuADv2.0-like datasets uses null answer \textbf{[CLS]} token probability
    to give a likelihood for a question to have an answer within the supporting passage
    ~\cite{sqaud2,bert}.
    This works by finding the difference between the null answer score of \textbf{[CLS]} token
    and the non-empty answer span with the highest score.
    $\varphi$ is a general span extractor that operates on a question $q$ and a passage $p$.
    \begin{equation}
        \operatorname{\gamma}(q,p) =  \operatorname{\varphi}(q, p)_{_{\textit{[CLS]}}} - \operatorname{\varphi}(q, p)_{_{\textit{MAX}}}
        \label{eq:threshold-b}
    \end{equation}

    Upon calculating scores for all samples, we proceed to normalize them into
    ${\operatorname{\bar{\gamma}}(q)}$ and then apply a threshold value $\zeta$ to
    determine if there is no answer. To identify unanswerable questions, we use a
    binary threshold function $\operatorname{\sigma}$,

    \begin{equation}
        \operatorname{\sigma}(q)= \mathds{1}_{{\operatorname{\bar{\gamma}}(q)} > \zeta}
        \label{eq:threshold-b2}
    \end{equation}

%

    \section{$\zeta$ Selection and $\zeta^\bigstar$}
    \label{appen:optimal}

    In this work, we defined $\zeta$ hyperparameter for zero-answer
    thresholding. This hyperparameter controls the proportion of samples that are
    considered to be zero-answer. Due to the small size of the dataset, we used a
    quantile method to set $\zeta$. This method marks a proportion of the samples
    according to the statistics of the dataset.
    Task~\textbf{B} is less sensitive to this parameter because almost 5\% of the samples are zero-answer.
    In contrast, task~\textbf{A} is highly sensitive to this parameter because of the larger proportion of zero-answer cases compared to task~\textbf{A}.
    Additionally, We are interested in finding the theoretical upperbound
    performance for  $\zeta$; this is addressed by \textbf{RQ3}.

    In Tables~\ref{tab:dev-a} and~\ref{tab:dev-b} we use $\bigstar$ accompanied by $\zeta$
    to refer to the optimal performance of the binary classification problem of has-answer vs. has-no-answer, as explained
    in Appendices~\ref{appen:threshold-a} and~\ref{appen:threshold-b}.
    Figure~\ref{fig:threshold} illustrates the thresholding effect against fine-tuned model performance for task~\textbf{A}; this answers \textbf{RQ3}.
    As we can see, the $\zeta$ hyperparameter can not be set arbitrarily.
    Instead, we can adjust it by considering the outcomes obtained from trained models on the training data.
    To find the optimal threshold $\zeta^\bigstar$ for both tasks, we implemented a greedy optimization algorithm for
    all possible levels of thresholds made by a given model; check the code for more details~\footnote{
        In both code bases, this is performed by function \textit{find\_best\_thresh}.
        You may find this function under \textit{metrics} directory in \textit{compute\_score\_qrcd.py} and
        \textit{helpers.py} scripts for tasks~\textbf{A}~and~\textbf{B}, respectively.
    }.



\end{document}